%% file: ICRA_2026_Sebnem.tex
\documentclass[letterpaper, 10 pt, conference]{IEEEconf} 
\IEEEoverridecommandlockouts
\usepackage{cite}
\usepackage{amsmath,amssymb,amsfonts}
\usepackage{algorithm}
\usepackage[noend]{algpseudocode}
\usepackage{graphicx}
\usepackage{multirow} 
\usepackage{textcomp}
\usepackage{tabularray}
\usepackage{tabularx}
\usepackage{graphicx}  
\usepackage{booktabs}
\usepackage{soul,color}
\usepackage[table]{xcolor}
\usepackage[inkscapeformat=pdf]{svg}
\usepackage[caption=false,font=footnotesize]{subfig}

\def\BibTeX{{\rm B\kern-.05em{\sc i\kern-.025em b}\kern-.08em
    T\kern-.1667em\lower.7ex\hbox{E}\kern-.125emX}}

\usepackage{scalerel} 
\usepackage{tikz} 
\usetikzlibrary{svg.path} 

\definecolor{orcidlogocol}{HTML}{A6CE39}
\tikzset{
  orcidlogo/.pic={
    \fill[orcidlogocol] svg{M256,128c0,70.7-57.3,128-128,128C57.3,256,0,198.7,0,128C0,57.3,57.3,0,128,0C198.7,0,256,57.3,256,128z};
    \fill[white] svg{M86.3,186.2H70.9V79.1h15.4v48.4V186.2z}
                 svg{M108.9,79.1h41.6c39.6,0,57,28.3,57,53.6c0,27.5-21.5,53.6-56.8,53.6h-41.8V79.1z M124.3,172.4h24.5c34.9,0,42.9-26.5,42.9-39.7c0-21.5-13.7-39.7-43.7-39.7h-23.7V172.4z}
                 svg{M88.7,56.8c0,5.5-4.5,10.1-10.1,10.1c-5.6,0-10.1-4.6-10.1-10.1c0-5.6,4.5-10.1,10.1-10.1C84.2,46.7,88.7,51.3,88.7,56.8z};
  }
}

\newcommand\orcidicon[1]{\href{https://orcid.org/#1}{\mbox{\scalerel*{
\begin{tikzpicture}[yscale=-1,transform shape]
\pic{orcidlogo};
\end{tikzpicture}
}{|}}}}

\usepackage[hidelinks]{hyperref}
\begin{document}
\title{\LARGE \bf Edged USLAM: Edge-Aware Event-Based SLAM with Learning-Based Depth Priors}


\author{Şebnem Sarıözkan \textsuperscript{\orcidicon{0009-0003-2789-0648}}, 
Hürkan Şahin \textsuperscript{\orcidicon{0009-0008-7920-5872}},
Olaya Álvarez-Tuñón \textsuperscript{\orcidicon{0000-0003-3581-9481}},
and Erdal Kayacan \textsuperscript{\orcidicon{0000-0002-7143-8777}}%
\thanks{*This work was supported by the Horizon Europe Grant Agreement No. 101136056.}
\thanks{Ş. Sarıözkan, H. Şahin, and E. Kayacan are with the Automatic Control Group (RAT), Paderborn University, 33098 Paderborn, Germany (e-mail: \{sariozka, hursah\}@mail.uni-paderborn.de and erdal.kayacan@uni-paderborn.de). 
O. Álvarez-Tuñón is with EIVA A/S, and IT University of Copenhagen (ITU), Denmark (e-mail: oltu@itu.dk).
}}


\maketitle

\input{sections/00_Abstract}
\input{sections/01_Intro}
\input{sections/02_RelatedWork}
\input{sections/03_Methodology}

\input{sections/04_Experiments}

\input{sections/05_Conclusions}



\bibliographystyle{IEEEtran}
\bibliography{References}

\end{document}

%% file: sections/00_Abstract.tex
\begin{abstract}



Conventional visual simultaneous localization and mapping (SLAM) algorithms often fail under rapid motion, low illumination, or abrupt lighting transitions due to motion blur and limited dynamic range. Event cameras mitigate these issues with high temporal resolution and high dynamic range (HDR), but their sparse, asynchronous outputs complicate feature extraction and integration with other sensors, e.g., inertial measurement units (IMUs) and standard cameras. We present Edged USLAM, a hybrid visual–inertial system that extends Ultimate SLAM (USLAM) with an edge-aware front-end and a lightweight depth module. The front-end enhances event frames for robust feature tracking and nonlinear motion compensation, while the depth module provides coarse, region-of-interest (ROI)-based scene depth to improve motion compensation and scale consistency. Evaluations across public benchmarks and real-world unmanned air vehicle (UAV) flights demonstrate that performance varies significantly by scenario. For instance, event-only methods like point–line event-based visual–inertial odometry (PL-EVIO) or learning-based pipelines such as deep event-based visual odometry (DEVO) excel in highly aggressive or extreme HDR conditions. In contrast, Edged USLAM provides superior stability and minimal drift in slow or structured trajectories, ensuring consistently accurate localization on real flights under challenging illumination. These findings highlight the complementary strengths of event-only, learning-based, and hybrid approaches, while positioning Edged USLAM as a robust solution for diverse aerial navigation tasks.



\end{abstract}

%% file: sections/01_Intro.tex
\section{Introduction}

Simultaneous localization and
mapping (SLAM) is a cornerstone of autonomous navigation, enabling robots to build a map of their surroundings while estimating their own motion within it. Despite its success in ground vehicles \cite{montemerlo2008darpa}, augmented reality/virtual reality devices \cite{klein2007ptam}, and service robots \cite{stachniss2005robot}, the demand for robust SLAM is particularly critical in aerial robotics, where drones operate in hazardous or inaccessible environments \cite{pham2022deep}. Examples include collapsed buildings after earthquakes \cite{kleiner2007rescue}, confined ship compartments during recycling and inspection tasks \cite{2021ship}, industrial facilities with toxic atmospheres 
or underground tunnels and mines \cite{tunnel2020}. In these scenarios, global positioning system (GPS) signals are often unavailable, making onboard, real-time state estimation essential for safe and reliable operation.

Conventional vision-based SLAM algorithms face significant limitations. Standard cameras, the primary sensing modality in most pipelines, struggle with fast motion \cite{andersen2022event} or sudden illumination changes due to motion blur, limited dynamic range, and high latency \cite{mueggler2020eventhdr}. These shortcomings lead to unreliable localization in challenging environments where drones are most critical, such as dark, cluttered, or dynamic settings like collapsed structures or industrial sites. Such limitations highlight the need for alternative sensing approaches to ensure robust navigation in these demanding scenarios \cite{alvarez2020monocular}.

\begin{figure}[!t]
\centering
\includegraphics[width=0.9\linewidth]{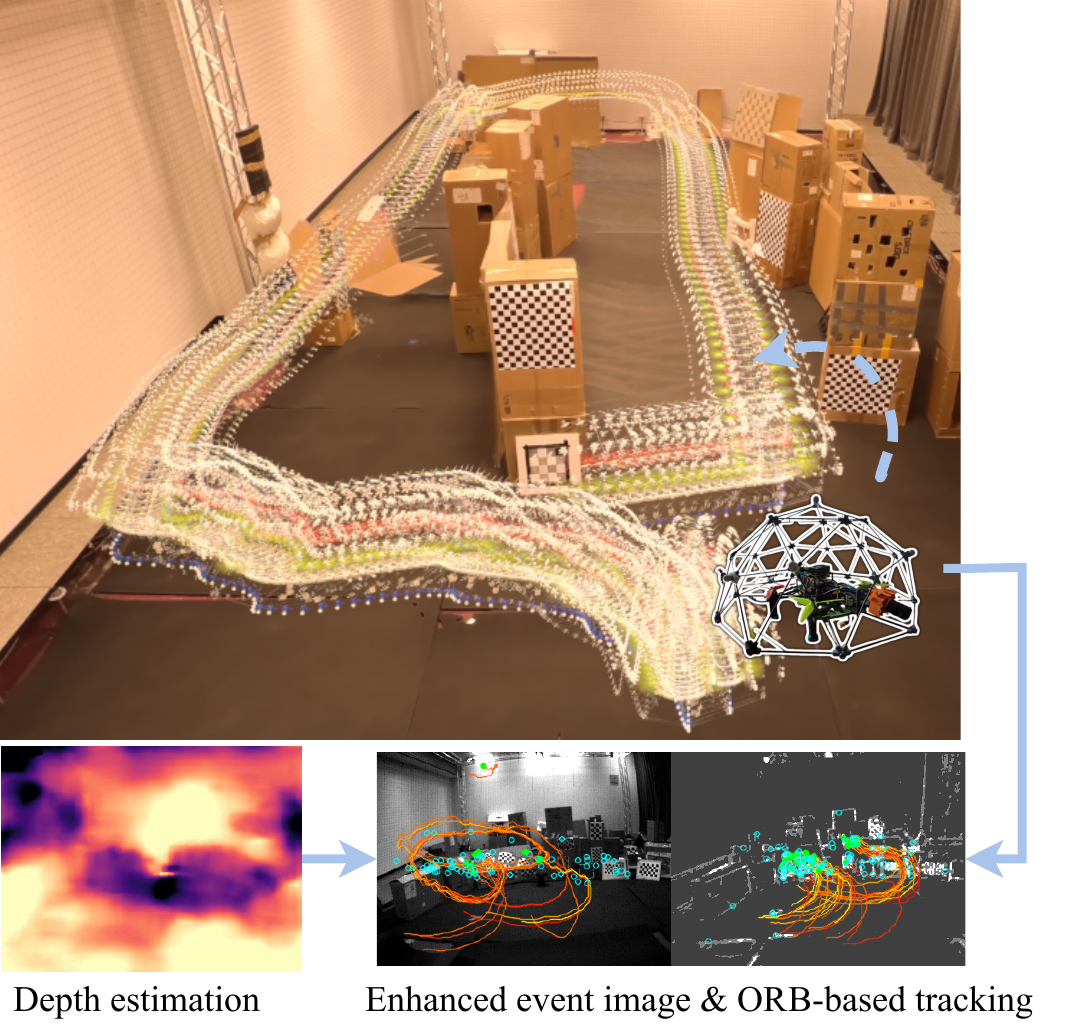}
\caption{Overview of the proposed Edged USLAM framework. The UAV (right) equipped with a protective cage, event camera, and depth module navigates through a cluttered indoor environment (top). The pipeline integrates learning-based depth estimation (bottom left) with enhanced event frames for robust ORB-based tracking (bottom center-right), enabling accurate 3D trajectory reconstruction (top).}
\label{fig:intro}
\end{figure}

To address the mentioned limitations, event cameras provide a biologically inspired sensing alternative. Unlike conventional cameras, which capture frames at fixed intervals, event cameras asynchronously detect per-pixel brightness changes with microsecond latency and high dynamic range \cite{gallego2020event}. Consequently, event cameras are ideal for fast motion, HDR, and low-light conditions. However, their sparse and asynchronous output is fundamentally different from conventional RGB images, posing unique challenges for SLAM. In particular, the lack of dense grid-structured data leads to a representation mismatch, hindering the use of standard feature extraction and matching techniques \cite{rebecq2016evo,gallego2020event}. Additionally, the microsecond-level timestamps of events introduce asynchronous fusion difficulties when integrating with frame-based sensors or IMUs, requiring specialized synchronization and motion-compensation strategies as seen in \cite{vidal2018ultimate,gehrig2021dsec}.

This work presents Edged USLAM\footnote{Project page at: \href{https://sebnem-byte.github.io/edged-uslam.github.io/}{https://sebnem-byte.github.io/edged-uslam.github.io/}} as illustrated in Fig.~\ref{fig:intro}, a hybrid visual--inertial SLAM system that extends USLAM \cite{vidal2018ultimate} by introducing a lightweight, edge-aware front-end tailored for event data. The primary motivation is to reduce reliance on conventional RGB frames by enhancing the quality of event imagery itself, enabling robust localization even when standard intensity images are unavailable. Our method combines motion-compensated event frames with efficient edge filtering and grid-based oriented FAST and rotated BRIEF (ORB) feature tracking, improving robustness in low-light, high-speed, and low-featured scenes. Unlike prior methods that degrade in low-gradient or low-event conditions \cite{Alexander, niu2025esvo2}, the proposed pipeline enhances landmark consistency through stronger spatial saliency.

The proposed algorithm, Edged USLAM,  is validated on public benchmarks and real-time UAV flights with motion capture ground truth. We introduce a new aerial dataset that features diverse motion and illumination conditions for a realistic evaluation.  The experimental results presented show superior localization performance compared to ORB-SLAM3 and USLAM, achieving robustness in dark environments and during aggressive maneuvers.

The contributions of this paper are as follows:
\begin{itemize}
    \item A novel edge-aware front-end for event-based visual--inertial SLAM is designed, improving robustness under low-light and fast-motion conditions.
    \item Motion-compensated event frames are integrated with lightweight edge filtering, which is enhanced by a depth estimation module and grid-based ORB tracking to enhance landmark consistency.
    \item Real-time performance is demonstrated on a drone platform equipped with onboard computation, showing resilience in cluttered indoor environments.
    \item A new aerial dataset with synchronized events, IMU, and ground truth trajectories is provided, enabling future benchmarking in GPS-denied scenarios.
\end{itemize}

The remainder of the paper is organized as follows. Section~II reviews related work on event-based odometry and SLAM. Section III presents the Edged USLAM framework, detailing motion compensation, edge-aware feature extraction, and depth prior integration.  Section~IV describes the experimental setup and datasets, followed by quantitative and qualitative evaluations in Section~V. Finally, Section~VI concludes with a discussion of limitations and future research directions.

%% file: sections/02_RelatedWork.tex
\section{Related Work}
Visual SLAM has steadily evolved, beginning with filter-based monocular systems like MonoSLAM~\cite{davison2007monoslam} and advancing to modern multi-sensor frameworks such as ORB-SLAM3~\cite{campos2021orb} and VINS-Mono~\cite{qin2018vins}. Existing approaches can be broadly categorized into three paradigms: feature-based, direct, and learning-based methods. Feature-based pipelines extract sparse landmarks and remain the most widely adopted due to their robustness, computational efficiency, and suitability for real-time robotics~\cite{cadena2016past, barros2023survey}. Their main weakness lies in the reliance on  frame-based cameras, which often degrade under motion blur and low-light conditions. Direct methods such as LSD-SLAM~\cite{engel2014lsd} and DSO~\cite{engel2018direct} exploit pixel-level photometric consistency, enabling dense reconstructions without explicit feature extraction. However, they remain highly sensitive to changes in illumination and camera calibration errors. Learning-based methods, including VINet~\cite{clark2017vinet} and CNN-SLAM~\cite{tateno2017cnnslam}, leverage neural networks to infer depth or poses directly from raw images. While they offer strong adaptability and scene understanding, their performance is hindered by generalization issues, data requirements, and limited interpretability~\cite{alvarez2024loss}. Given these limitations across all paradigms, event-based vision has emerged as a promising alternative, offering high temporal resolution, low latency, and exceptional dynamic range, making it particularly well-suited for fast motion and challenging illumination conditions. \\
\indent Early efforts such as USLAM~\cite{vidal2018ultimate} marked a turning point by tightly coupling asynchronous events, standard frames, and IMU measurements within a single optimization pipeline. By generating motion-compensated event frames using depth from tracked landmarks, it enabled parallel feature tracking on both modalities and robust performance under aggressive motion. Building on this, PL-EVIO~\cite{PLEVIO} incorporated line features from event time surfaces to better exploit structured environments like corridors, while ESVO~\cite{gehrig2021esvo} departed from feature-based tracking and proposed a direct stereo formulation. The latter eliminated explicit event matching and enabled semi-dense reconstructions even in extremely low-light settings. More recently, EVI-SAM~\cite{zhang2022evisam} combined feature-based and direct residuals in a sliding-window framework, achieving accurate 6-DoF tracking and dense mapping by jointly exploiting event maps, time surfaces, and inertial cues. \\
\indent A complementary direction is learning-based event SLAM. Methods such as DEIO~\cite{GWPHKU_DEIO} and DEVO~\cite{DEVO} integrate neural depth estimation into the odometry pipeline, improving metric scale recovery and robustness in texture-poor or repetitive environments. This trend highlights how depth priors can act as stabilizing cues to compensate for the sparsity of event data.  In this trajectory, our Edged USLAM is positioned as a lightweight yet robust system. While not fully learning-based, it introduces two key ideas: an edge-aware front-end to enhance motion-compensated event frames, and a compact depth prior module that reduces scale drift. Together, these design choices improve landmark consistency and enable reliable tracking in visually degraded, GPS-denied environments.\\ 

%% file: sections/03_Methodology.tex
\section{Methodology}
\begin{figure}[!t]
\centering
\includegraphics[width=1.0\linewidth]{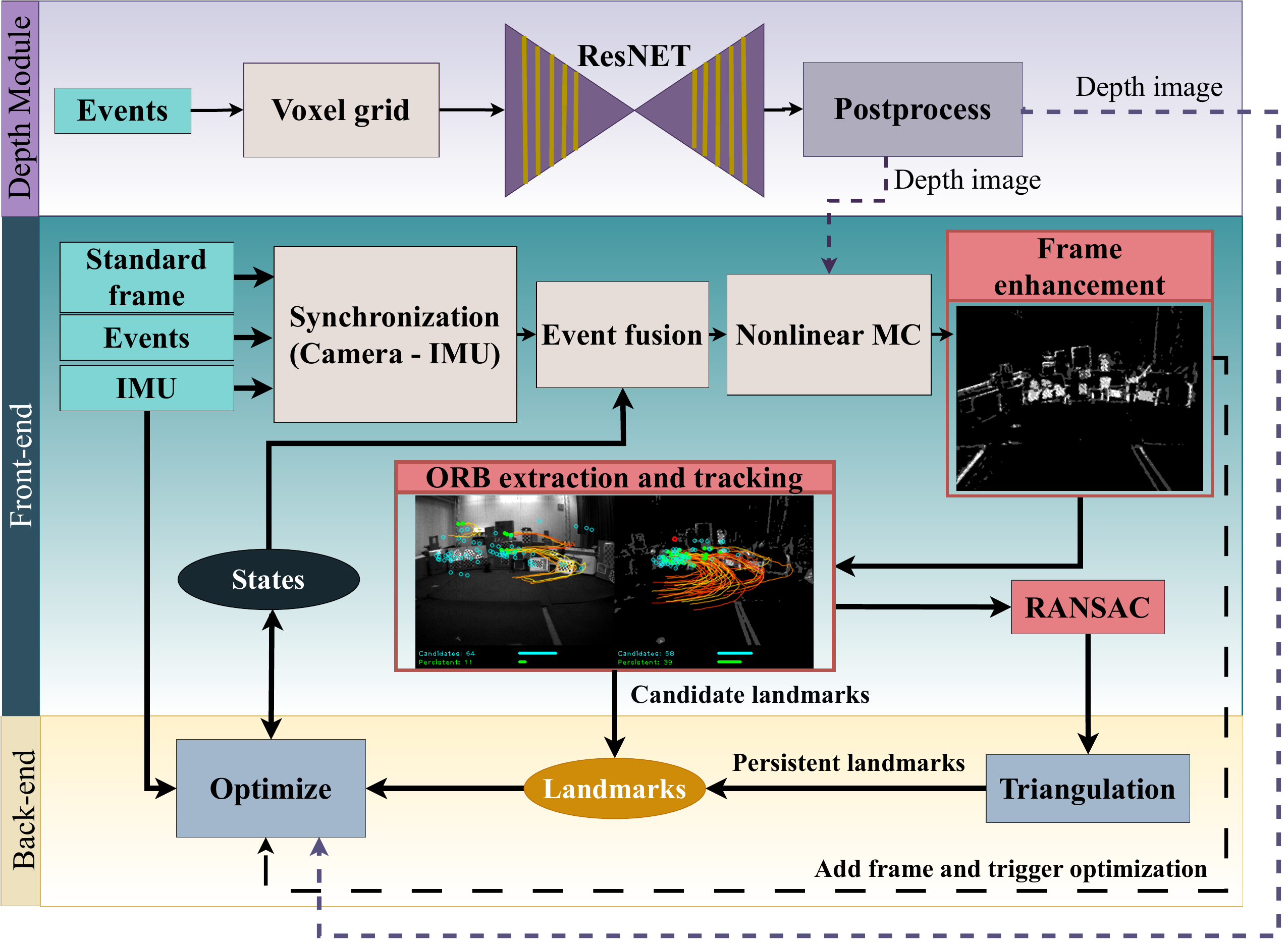 }
\caption{Overview of the proposed Edged USLAM methodology. 
Events, standard frames, and IMU data are synchronized and fused with nonlinear motion compensation, followed by frame enhancement and ORB-based feature tracking. 
Landmarks are refined via RANSAC and triangulation, while depth priors from the ResNet module support consistent state estimation and optimization in the back-end.}
\label{fig:Edged_USLAM}
\end{figure}

The proposed Edged USLAM architecture (Fig.~\ref{fig:Edged_USLAM}) builds upon the original Ultimate SLAM framework \cite{vidal2018ultimate}, which fuses events, standard frames, and IMU data in a tightly coupled pipeline. We retain the synchronization strategy and augmented event packet generation, but enhance both the motion model and event preprocessing. Events are warped using nonlinear pose interpolation with SE(3) transformations. Warped events are accumulated into sharper frames and enhanced via contrast and edge filters before ORB-based keypoints 
are extracted. In parallel, the depth module predicts dense event-based depth maps, from which scene-level priors are extracted. These priors support motion compensation in the front-end and are fused as soft constraints in the back-end optimization, improving scale consistency while maintaining real-time performance.

\subsection{Nonlinear motion compensation}
\label{sec:NMC}
Event cameras output asynchronous brightness changes, which are grouped into overlapping event packets synchronized with IMU data. Let $N$ denote the number of events in a packet and $i \in \{1,\dots,N\}$ be the event index. 
The timestamp of index $i$ is denoted by $t_i$. To compensate for motion, the camera pose at each event timestamp $t_i$ is obtained by interpolating between the start and end poses on $\mathrm{SE}(3)$:
\begin{equation}
\mathbf{T}_i = \mathbf{T}_0 \cdot \exp\!\big(\alpha_i \cdot \log(\mathbf{T}_1 \mathbf{T}_0^{-1})\big), 
\quad \alpha_i = \tfrac{t_i - t_0}{t_1 - t_0},
\end{equation}
where $\mathbf{T}_0, \mathbf{T}_1 \in \mathrm{SE}(3)$ denote the camera poses at the start and end of the event packet, and $\alpha_i$ is the normalized interpolation factor at time $t_i$.  

Each event is represented as $e_i=(\mathbf{x}_i,t_i,p_i)$, where $\mathbf{x}_i=(x_i,y_i)$ is the pixel location, $t_i$ the timestamp, and $p_i \in \{-1,+1\}$ the polarity. With an estimated depth $d_i$ and camera intrinsic matrix $\mathbf{K}$, the event is lifted to 3D, warped with $\mathbf{T}_i$, and reprojected to obtain the motion-compensated pixel:  
\begin{equation}
\hat{\mathbf{x}}_i 
= \pi\!\left(\mathbf{T}_i \, d_i \, \mathbf{K}^{-1}[x_i,y_i,1]^\top \right) 
- \Delta \mathbf{x}_{\text{last}},
\end{equation} where $\pi(\cdot)$ denotes the camera projection function and $\Delta \mathbf{x}_{\text{last}}$ is the previous alignment correction.  

This depth-aware motion compensation suppresses temporal smearing and yields sharper event frames, enabling more reliable feature extraction and tracking. The nonlinear warping in $\mathrm{SE}(3)$ is essential to accurately compensate rotational and translational motion, ensuring geometrically consistent reprojection.

\begin{figure}[!t]
\centering
\begin{tabular}{cc}
\subfloat[Motion compensation output]{%
  \includegraphics[width=0.45\columnwidth]{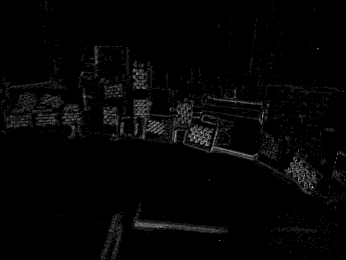}} &
\subfloat[CLAHE enhancement]{%
  \includegraphics[width=0.45\columnwidth]{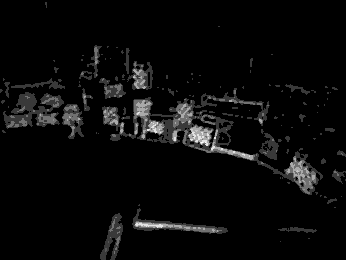}} \\

\subfloat[Canny edge detection]{%
  \includegraphics[width=0.45\columnwidth]{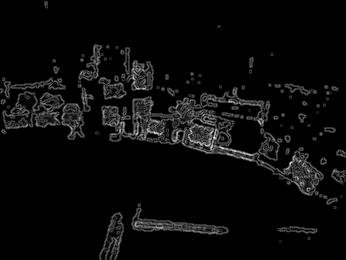}} &
\subfloat[Laplacian edge detection]{%
  \includegraphics[width=0.45\columnwidth]{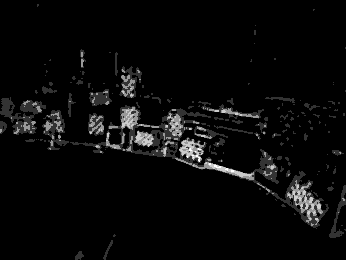}}
\end{tabular}
\caption{Comparison of event frame enhancement methods. 
(a) Motion compensation sharpens asynchronous event frames into a more stable representation. 
(b) CLAHE improves local contrast under varying illumination. 
(c) Canny edge detection extracts fine-grained structural contours. 
(d) Laplacian filtering highlights high-frequency edges and textures.}
\label{fig:event_enhancements}
\end{figure}

\subsection{Event image enhancement}

Event cameras produce temporally precise but spatially sparse and noisy data. 
To enable reliable feature extraction, we introduce an edge-enhanced preprocessing pipeline after motion compensation. 
First, a Gaussian blur reduces isolated noise, followed by contrast limited adaptive histogram equalization (CLAHE) to improve local contrast. 
An optional sharpening step further refines structural details. 
Subsequently, edge detectors (Canny, Laplacian, or Sobel) extract contours, with morphological erosion used when finer localization is required. 
The resulting edge map is blended with the original motion-compensated frame, yielding sharper and more structured event images. 
This representation improves the repeatability and localization of ORB keypoints, particularly in low-contrast or low-motion conditions.

\begin{algorithm}[!t]
\caption{Event Frame Enhancement}
\label{alg:enhance_event_frame}
\begin{algorithmic}[1]
\Function{EnhanceEventFrame}{$I_{\text{event}}$}
    \State $I_{\text{blur}} \gets \text{GaussianBlur}(I_{\text{event}}, \sigma)$
    \State $I_{\text{CLAHE}} \gets \text{CLAHE}(I_{\text{blur}})$
    \If {contrast\_sharpening is enabled}
        \State $I_{\text{enh}} \gets I_{\text{CLAHE}} + \lambda \cdot (I_{\text{CLAHE}} - I_{\text{blur}})$
    \Else
        \State $I_{\text{enh}} \gets I_{\text{CLAHE}}$
    \EndIf
    \If {method == ``Canny''}
        \State $I_{\text{edge}} \gets \text{Canny}(I_{\text{enh}})$
    \ElsIf {method == ``Laplacian''}
        \State $I_{\text{edge}} \gets \text{Laplacian}(I_{\text{enh}})$
    \ElsIf {method == ``Sobel''}
        \State $I_{\text{edge}} \gets \text{Sobel}(I_{\text{enh}})$
    \EndIf
    \If {edge\_thinning is enabled}
        \State $I_{\text{edge}} \gets \text{Erode}(I_{\text{edge}}, B)$
    \EndIf
    \State \Return $\alpha \cdot I_{\text{event}} + \beta \cdot I_{\text{edge}}$
\EndFunction
\end{algorithmic}
\end{algorithm}

The complete pipeline is outlined in Algorithm~\ref{alg:enhance_event_frame} and Fig.~\ref{fig:event_enhancements} 
shows an illustrative example of the enhancement stages, and compares the outcomes of different edge extraction strategies.

\subsection{Feature extraction and tracking}
Unlike  USLAM, which relies on FAST features, and other event-based 
methods that often employ conventional keypoint detectors, our system adopts 
a grid-based ORB strategy to enforce spatial uniformity. By dividing the 
event frame into an $R \times C$ grid, where $R$ and $C$ denote the number 
of grid rows and columns, and retaining the strongest keypoint per cell, 
the method mitigates feature clustering in textured regions and ensures more 
balanced geometric coverage, leading to improved robustness in low-texture 
and high-speed scenarios. 
Let $\mathcal{K} = \{k_1, \dots, k_N\}$ be the set of all detected ORB keypoints 
in the image $I \in \mathbb{R}^{H \times W}$, where $H$ and $W$ denote the image height and width, and each keypoint 
$k_i = (\mathbf{x}_i, r_i)$ consists of a position $\mathbf{x}_i$ and a response $r_i$. 
We denote by $\mathcal{K}_{i,j}$ the subset of keypoints within grid cell $G_{i,j}$, 
and by $k^{*}_{i,j}$ the selected keypoint with the maximum response in that cell. 
The final selection $\mathcal{K}^*$ is then obtained as the union of all 
$k^{*}_{i,j}$ across the grid. 
\begin{equation}
k^{*}_{i,j} = \operatorname*{arg\,max}_{k \in \mathcal{K}_{i,j}} \mathrm{response}(k),
\qquad
\mathcal{K}^{*} = \bigcup_{i=1}^{R} \bigcup_{j=1}^{C} \{\,k^{*}_{i,j}\,\}.
\end{equation}
where $ k_{i,j} \subset \mathcal{K}$ denotes keypoints within grid cell $\mathcal{G}_{i,j}$.
Feature tracking is performed with a pyramidal KLT optical flow. 
Each landmark is projected into the current frame using the relative pose, 
and the tracker refines its position by minimizing the photometric error 
over a local patch.


\subsection{Depth module integration in Edged-USLAM}

\begin{figure}[!t]
\centering
\begin{tabular}{cccc}
\subfloat[Event input 1]{%
  \includegraphics[width=0.22\textwidth]{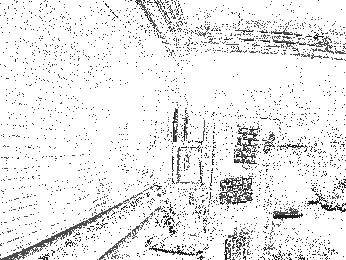}} &
\subfloat[Event input 2]{%
  \includegraphics[width=0.22\textwidth]{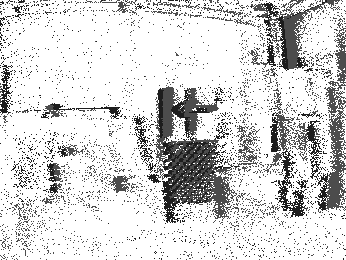}} \\

\subfloat[ Depth result 1]{%
  \includegraphics[width=0.22\textwidth]{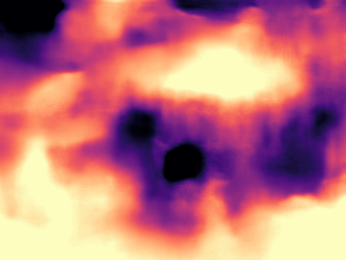}} &
\subfloat[ Depth result 2]{%
  \includegraphics[width=0.22\textwidth]{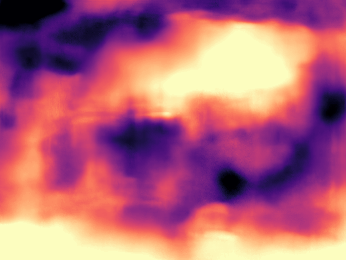}}
\end{tabular}
\caption{Real-time demonstration of event-based depth estimation. 
Top row: sample event inputs (a–b). 
Bottom row: corresponding coarse depth estimation results (c–d) computed by a lightweight model for rapid inference, intended not for accurate reconstruction but for rapid inference of approximate object distances.}
\label{fig:event_depth_demo}
\end{figure}

To enable real-time metric scale recovery, we integrate the 
learning-based event-to-depth model 
(\href{https://github.com/uzh-rpg/rpg_e2depth}{e2depth})~\cite{HidalgoCarrio2020Learning} 
into the Edged USLAM framework. The network is used in a lightweight configuration, taking event-voxel grids as input and providing a scene-depth estimate. For deployment, we implement a simplified ROS node that omits non-critical modules, makes the recurrent state optional, and applies fast voxel cropping and normalization, reducing latency for online inference. Specifically, we compute the mean depth $\bar{d}_t$ within a central ROI and stabilize it with exponential smoothing:
\begin{equation}
    d_t = \alpha \bar{d}_t + (1-\alpha) d_{t-1}, 
    \quad d_t \in (d_{\min}, d_{\max}),
\end{equation}
where $\alpha$ is a smoothing factor and $(d_{\min}, d_{\max})$ are 
validity bounds. The resulting $d_t$ is published as a scene-depth prior and fused with landmark-based triangulation inside the front-end.


In the back-end, the ROI-based scene depth $d_t$ is converted to inverse depth $\rho^{\text{ROI}}_t$ and used as a soft prior in the factor graph:

\begin{equation}
E_{\text{depth}} = \left\|\, \bar{\rho}_t - s\,\rho^{\text{ROI}}_t \,\right\|^2_{\Sigma_t},
\end{equation}
where $\bar{\rho}_t$ is the optimized average inverse depth in the current 
keyframe, $s$ is a global scale parameter, and $\Sigma_t$ encodes confidence. 

Figure~\ref{fig:event_depth_demo} illustrates the intermediate depth maps. Although coarse, these lightweight predictions are geometry-aware and enable real-time motion compensation and scale consistency.



%% file: sections/04_Experiments.tex
\section{Experiments}

The proposed SLAM system was evaluated in real-world conditions using a quadrotor platform equipped with a PX4-based Pixhawk 6X flight controller. The onboard perception stack runs on an NVIDIA Jetson Orin NX mission computer, which processes data from a DAVIS 346 event camera for event-based visual–inertial odometry. Localization performance was assessed by comparing the SLAM-estimated poses against ground truth provided by a motion capture system, operating at high frequency with full 6-DoF accuracy.

\begin{figure}[!t]
    \centering
    \begin{minipage}{0.48\linewidth}
        \centering
        \subfloat[Front view]{%
        \includegraphics[width=\linewidth]{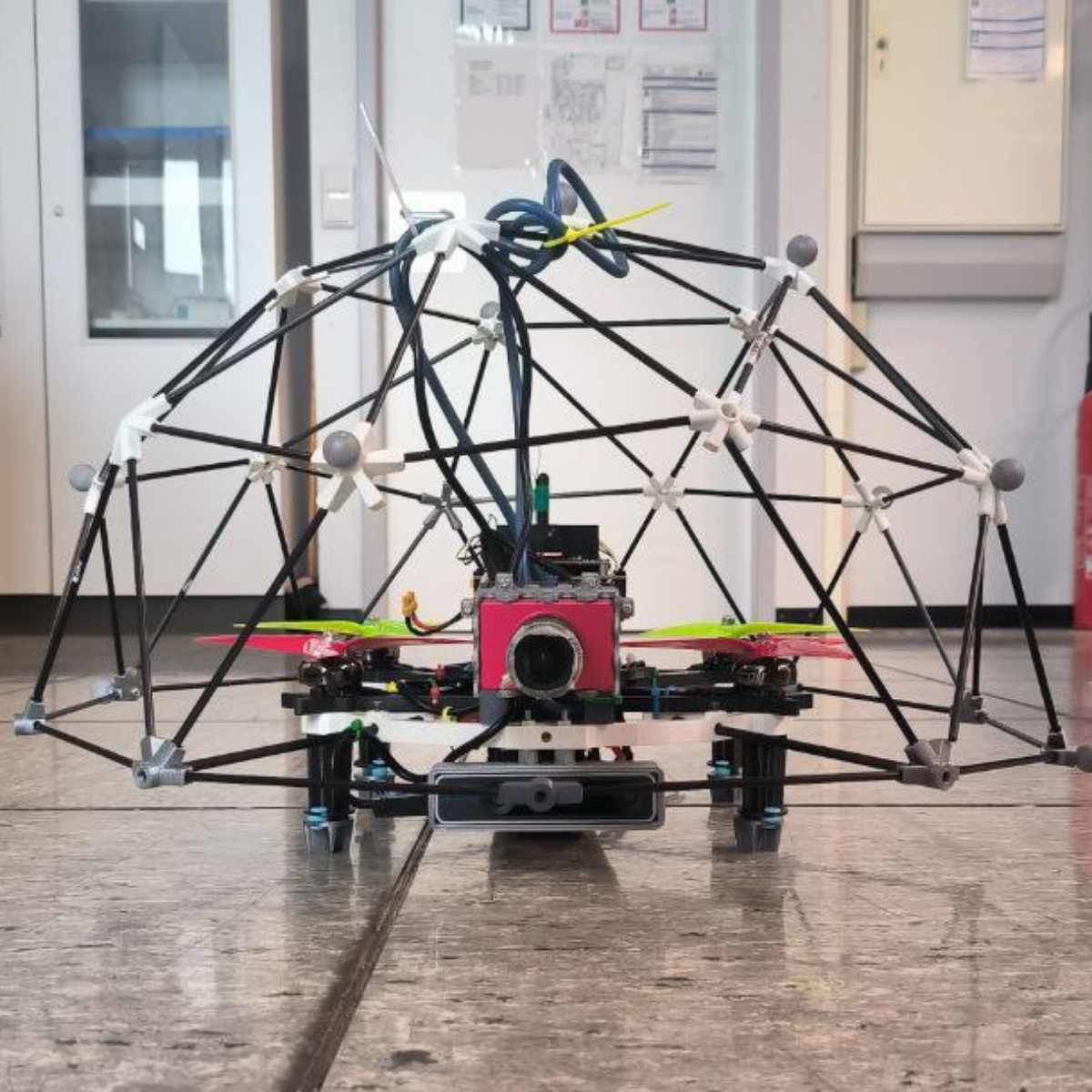} }
    \end{minipage}
    \hfill
    \begin{minipage}{0.48\linewidth}
        \centering
        \subfloat[Side view]{%
        \includegraphics[width=\linewidth]{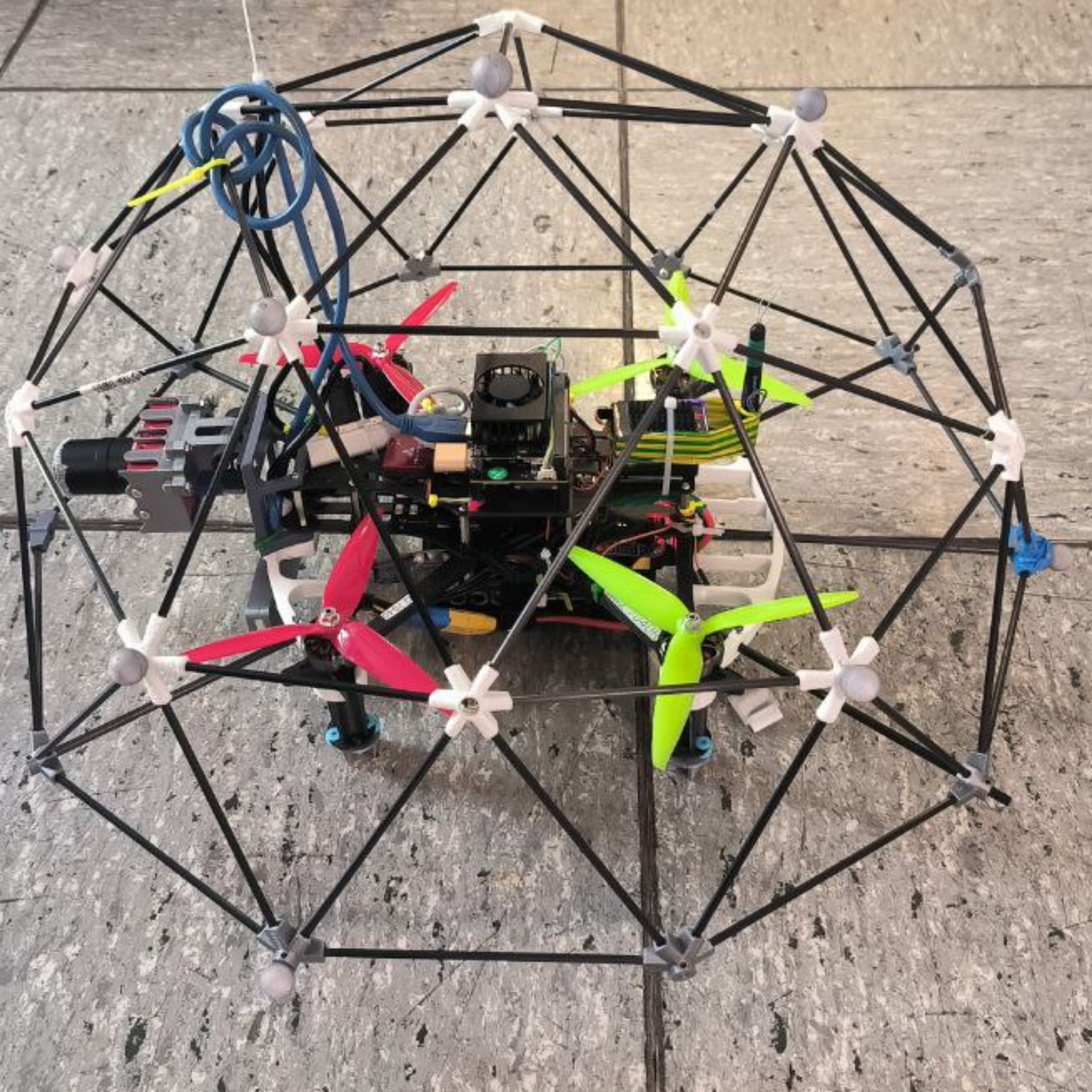} }
    \end{minipage}
     \caption[Test platform used for SLAM evaluation.]{TESTUDO an UAV platform equipped with a collision-tolerant protective cage and multi-sensor payload (event camera, tracking sensor, and IMU) for real-time SLAM evaluation in confined and hazardous environments}
    \label{fig:testudo}
\end{figure}

As shown in Fig.~\ref{fig:testudo}, the TESTUDO is designed as a collision-tolerant quadrotor for operation in confined and hazardous environments. The lightweight geodesic cage protects both the drone and its payload during impacts, allowing safe navigation in cluttered spaces.

\subsection{Evaluation on benchmark sequences}

For a detailed evaluation of our system, we considered sequences from multiple publicly available event-based datasets. 
Specifically, representative sequences from the Event-Camera Dataset and Simulator~\cite{mueggler2017ijrr} were used to cover challenging motion and lighting conditions, 
including aggressive 6-DoF motion, dynamic objects, and pure translational motion. 
In addition, sequences from the HKU Event-based VO/VIO/SLAM dataset~\cite{hku_event_slam} were included to validate performance in indoor and fast-motion UAV scenarios. 
We further tested on the recently introduced VECtor dataset~\cite{vector_dataset}, which provides large-scale and diverse recordings under real-world conditions. 
Together, these datasets offer broad coverage of motion dynamics, illumination changes, and scene complexity, enabling a thorough evaluation of robustness and generalization.


\begin{table}[!b]
\centering
\caption{Position RMSE (m) on the Event-Camera Dataset and Simulator dataset with event-based methods.}
\label{tab:eth_classical}
\renewcommand{\arraystretch}{1.2}
\setlength{\tabcolsep}{2pt} 
\resizebox{\columnwidth}{!}{
\begin{tabular}{lcccccc}
\toprule
\textbf{Sequence} & \textbf{Edged(ours)} & \textbf{EVI-SAM\cite{zhang2022evisam}} & \textbf{USLAM\cite{vidal2018ultimate}} & \textbf{PL-EVIO \cite{PLEVIO}} & \textbf{EKLT-VIO\cite{EKLTVIO}} \\ 
\midrule
boxes\_6dof           & \textbf{0.14} & 0.16 & 0.30 & 0.21  & 0.84 \\
boxes\_trans          & 0.16 & 0.11 & 0.27 & \textbf{0.06}  & 0.48 \\
hdr\_boxes            & 0.22 & 0.13 & 0.37 & \textbf{0.10}  & 0.46 \\
dynamics\_trans       & \textbf{0.10} & 0.30 & 0.18 & 0.24  & 0.40 \\
poster\_trans         & \textbf{0.07} & 0.34 & 0.12 & 0.54  & 0.35 \\
hdr\_poster           & 0.18 & 0.15 & 0.31 & \textbf{0.12}  & 0.65 \\
dynamic\_6DOF         & \textbf{0.12} & 0.27 & 0.19 & 0.48  & 0.79 \\
\midrule
\textbf{Average}      & \textbf{0.14} & 0.17 & 0.23 & 0.23 & 0.56 \\
\bottomrule
\end{tabular} }
\label{table:ETHresults}
\end{table}

\begin{figure}[!t]
\centering
\newcommand{\colw}{1\columnwidth} 

\begin{minipage}[t]{\colw}\centering
  \subfloat[X axis position]{%
    \includegraphics[width=\linewidth]{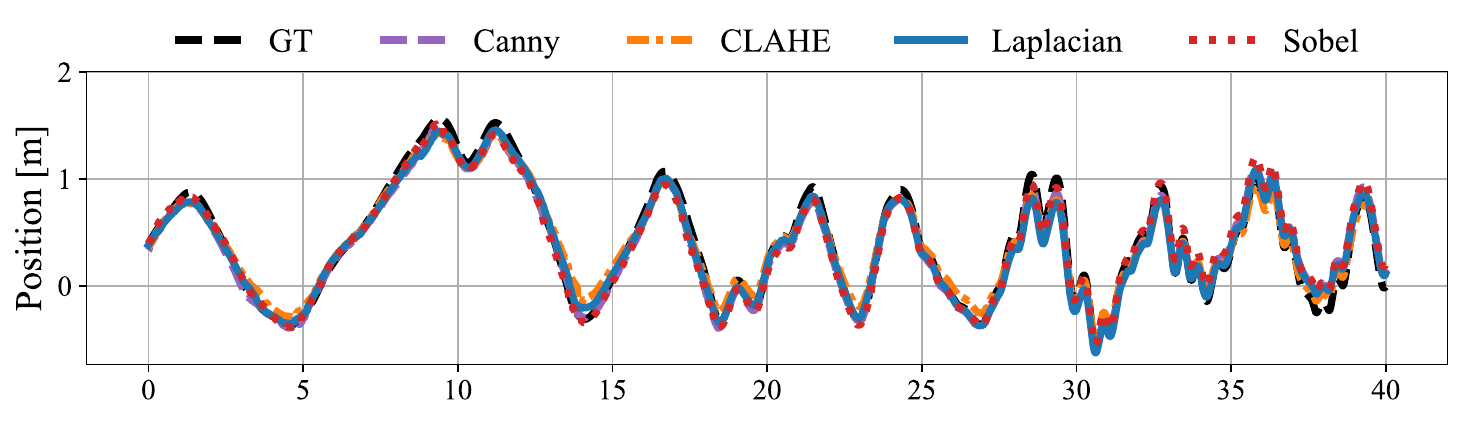}%
    \label{fig:slam_results_boxes:x}}
\end{minipage}

\vspace{0.6ex}

\begin{minipage}[t]{\colw}\centering
  \subfloat[Y axis position]{%
    \includegraphics[width=\linewidth]{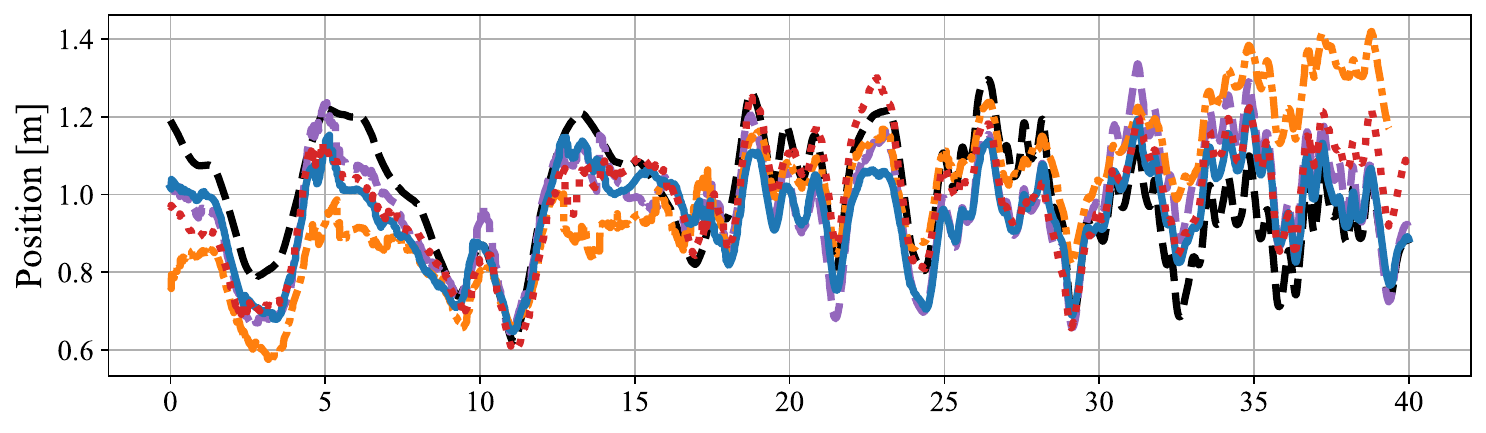}%
    \label{fig:slam_results_boxes:y}}
\end{minipage}

\vspace{0.6ex}

\begin{minipage}[t]{\colw}\centering
  \subfloat[Z axis position]{%
    \includegraphics[width=\linewidth]{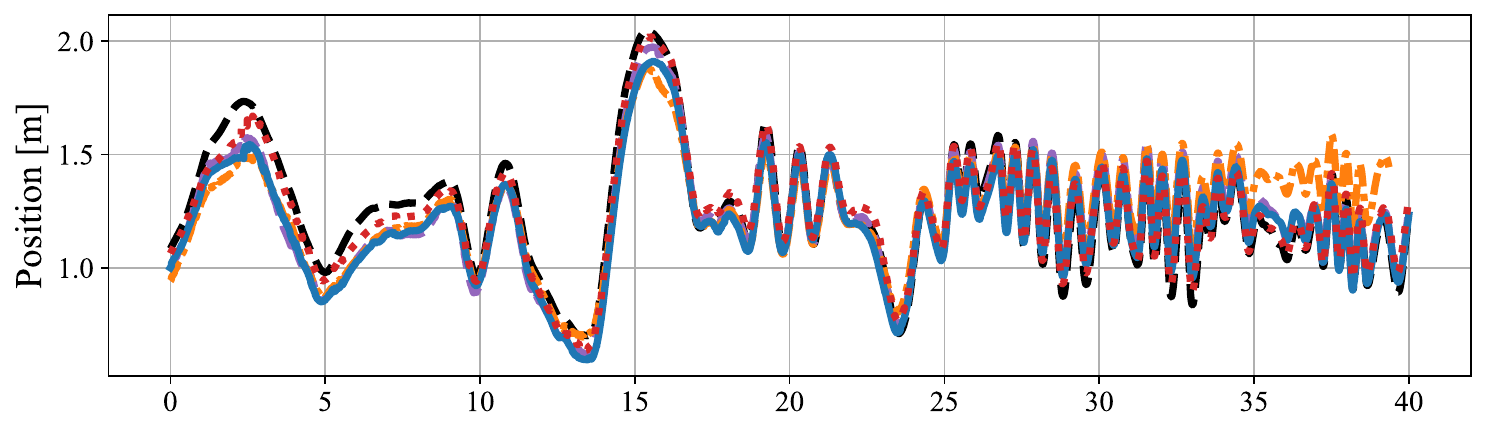}%
    \label{fig:slam_results_boxes:z}}
\end{minipage}

\vspace{0.6ex}

\begin{minipage}[t]{\colw}\centering
  \subfloat[APE error]{%
    \includegraphics[width=\linewidth]{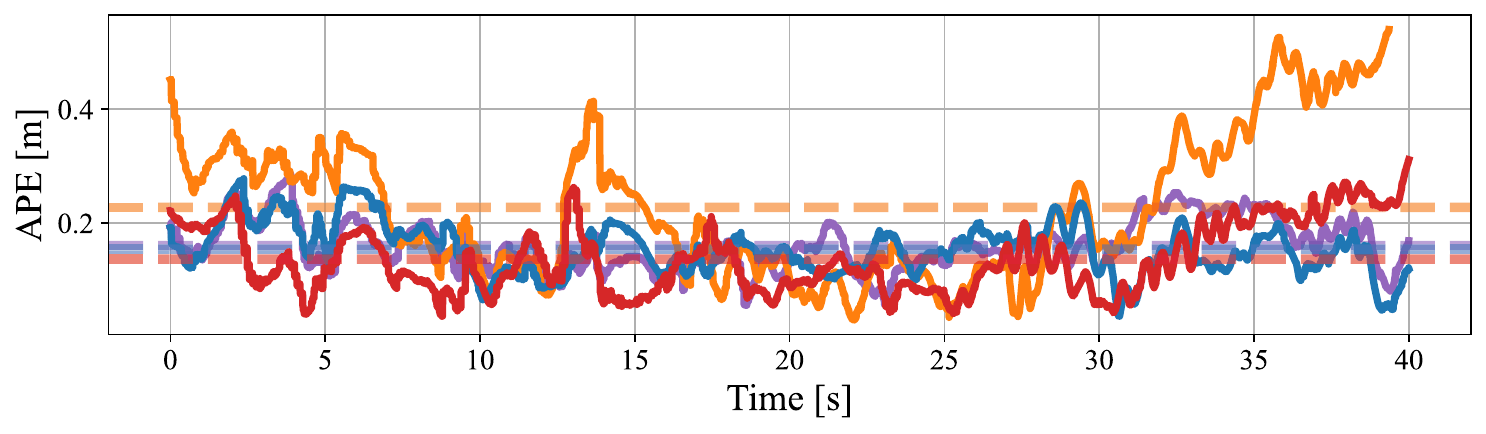}%
    \label{fig:slam_results_boxes:ape}}
\end{minipage}

\caption{Comparison of four event frame enhancement methods 
(Canny, CLAHE, Laplacian, Sobel) on the boxes\_6dof sequence. 
(a--c) Estimated X, Y, and Z axis positions compared with ground truth (black dashed). 
(d) Absolute position error (APE) over time. Sobel achieves the lowest mean error, 
followed by Canny and Laplacian, while CLAHE yields the highest errors. 
Mean APE values: Sobel is 0.143\ m, Canny is 0.167\ m, Laplacian is 0.175\ m, and CLAHE is 0.227\ m. 
}
\label{fig:slam_results_boxes}
\end{figure}

\begin{table*}[!t]
\centering
\caption{Position RMSE (m) on HKU and Vector datasets with classical and learning-based methods.}
\label{tab:hku_vector_all}
\renewcommand{\arraystretch}{1.2}
\setlength{\tabcolsep}{6pt} 
\begin{tabular}{llc|cccc|cc}
\toprule
 & & \multicolumn{1}{c|}{\textbf{Hybrid}}  & \multicolumn{4}{c|}{\textbf{Standard (Non-learning)}}  & \multicolumn{2}{c}{\textbf{Learning-based}} \\
\cmidrule(lr){3-7}\cmidrule(lr){8-9}
\textbf{Dataset} & \textbf{Sequence} & \textbf{Edged(ours)} & \textbf{EVI-SAM\cite{zhang2022evisam}} & \textbf{USLAM\cite{vidal2018ultimate}} & \textbf{PL-EVIO\cite{PLEVIO}} & \textbf{EVO\cite{rebecq2016evo}} & \textbf{DPVO\cite{DPVO}} & \textbf{DEVO\cite{DEVO}} \\
\midrule
\multirow{5}{*}{HKU}
  & agg\_tran      & 0.13 & 0.17 & 0.59 & 0.07 & \textit{failed} & 0.12 & \textbf{0.06} \\
  & agg\_rot       & 0.25 & 0.24 & 3.14 & 0.23 & \textit{failed} & 0.28 & \textbf{0.05} \\
  & agg\_walk      & \textbf{0.19} & 0.42 & 2.00 & 0.42 & \textit{failed} & \textit{failed} & 0.9 \\
  & hdr\_circle    & 0.49 & \textbf{0.13} & 1.32 & 0.14 & \textit{failed} & 1.19 & 0.39 \\
  & hdr\_slow      & 0.13 & 0.11 & 2.80 & 0.13 & \textit{failed} & 0.36 & \textbf{0.08} \\
\midrule
\multirow{5}{*}{VECtor}
  & corner\_slow   & \textbf{0.39} & 2.50 & 4.83 & 2.10 & 4.83 & \textit{failed}  & 0.59 \\
  & robot\_norm    & \textbf{0.14} & 0.67 & 1.81 & 0.68 & \textit{failed}  & 0.22      & 0.17 \\
  & robot\_fast    & 0.56 & 0.22 & 1.65 & 0.14 & \textit{failed}  & 0.78      & \textbf{0.13} \\
  & sofa\_fast     & 0.67 & 0.98 & 2.54 & \textbf{0.17} & \textit{failed}  & 0.60      & 0.38 \\
  & sofa\_norm     & 0.59 & \textbf{0.19} & 5.74 & \textbf{0.19} & \textit{failed}  & 0.22      & 0.22 \\
\midrule
\multicolumn{2}{l}{\textbf{Average}} & \textbf{0.35} & 0.56 & 2.64 & 0.42 & 0.47 & \ 0.47 & 0.29 \\
\bottomrule
\end{tabular}
\end{table*}

\begin{table*}[!t]
\centering
\caption{Absolute Pose Error (APE, m) and range (Min–Max) across motion categories and illumination conditions on our dataset.}
\label{tab:ape_motion_illum}
\setlength{\tabcolsep}{12pt}
\renewcommand{\arraystretch}{1.1}
\resizebox{\textwidth}{!}{%
\begin{tabular}{llccc ccc ccc}
\toprule
 & & \multicolumn{3}{c}{\textbf{Edged USLAM (ours)}} & \multicolumn{3}{c}{\textbf{ORB-SLAM3 \cite{campos2021orb}}} & \multicolumn{3}{c}{\textbf{USLAM\cite{vidal2018ultimate}}} \\
\cmidrule(lr){3-5}\cmidrule(lr){6-8}\cmidrule(lr){9-11}
\textbf{Category} & \textbf{Sequence} & \textbf{APE} & \textbf{Max} & \textbf{Min} & \textbf{APE} & \textbf{Max} & \textbf{Min} & \textbf{APE} & \textbf{Max} & \textbf{Min} \\
\midrule
\multirow{6}{*}{Motion}
 & circle                 & 0.26 & 0.64 & 0.02 & \textbf{0.21} & 2.1 & 0.01 & 0.48 & 1.1 & 0.1 \\
 & aggressive             & \textbf{0.19} & 0.62 & 0.015  & 0.57   & 1.75 & 0.05   & 0.35   & 0.75 & 0.12 \\
 & aggressive-lowlit      & \textbf{0.28} & 0.58 & 0.05 & \textit{failed} & -- & -- & 0.32 & 1.0 & 0.02 \\
 & square                 & 0.32 & 0.72 & 0.06& \textbf{0.19} & 0.96 & 0.035 & 0.71 & 2.1 & 0.10 \\
 & line                   & 0.21 & 0.66 & 0.03 & 0.21 & 0.61 & 0.036 & 0.47 & 1.5 & 0.09 \\
 & turned                 & \textbf{0.38} & 0.86 & 0.10 & \textit{failed} & -- & -- & 0.55 & 1.35 & 0.10 \\
\midrule
\multirow{6}{*}{Illumination}
 & low-lit                & \textbf{0.34} & 0.72 & 0.014 & \textit{failed} & -- & -- & 0.85 & 1.6 & 0.15 \\
 & 30\% lit               & \textbf{0.54} & 1.10 & 0.12 & \textit{failed} & -- & -- & 0.58 & 2.1 & 0.18 \\
 & 60\% lit               & \textbf{0.45} & 0.96 & 0.08 & 1.7 & 3.8 & 0.06 & 0.54 & 1.17 & 0.06 \\
 & dynamic HDR            & \textbf{0.76} & 1.56 & 0.11 & \textit{failed} & -- & -- & 1.49 & 3.2 & 0.25 \\
 & constant HDR           & \textbf{0.52} & 1.45 & 0.09 & 1.45 & 3.6 & 0.21 & 0.71 & 1.69 & 0.2 \\
\midrule
\multicolumn{2}{l}{\textbf{Average}} 
 & \textbf{0.38} &  &  & 0.72 &  &  & 0.64 &  &  \\
\bottomrule
\end{tabular}%
}
\end{table*}

The position RMSE results on the Event-Camera Dataset and Simulator are reported in Table~\ref{table:ETHresults}.
Edged USLAM achieves the lowest average error (0.14 m) and provides the best accuracy in four out of seven sequences. Compared to USLAM~\cite{vidal2018ultimate}, it consistently reduces drift, with the largest improvements observed in dynamic sequences. The improvement indicates that adding edge-aware event processing and depth cues provides stronger geometric consistency than USLAM’s original fusion of frames, events, and inertial data.  

Compared to PL-EVIO~\cite{PLEVIO}, which is purely event-driven, our method performs better in dynamic and cluttered environments, where the absence of appearance and depth information causes larger drift for PL-EVIO. EVI-SAM~\cite{zhang2022evisam} benefits from semantic priors and yields stable results in static scenes, but remains sensitive to fast motion and moving objects, where Edged USLAM maintains more reliable tracking. Finally, EKLT-VIO~\cite{EKLTVIO} shows the weakest performance overall due to the limited geometric constraints of event-based KLT tracking, while our method reduces drift substantially across all sequences.

To assess the impact of the edge-enhancement module, we compared four filtering strategies—Canny, CLAHE, Laplacian, and Sobel—on the boxes\_6dof sequence. 
As shown in Figure~\ref{fig:slam_results_boxes}, CLAHE, while primarily a local contrast enhancement technique rather than an edge detector, increases the number of detectable features by amplifying intensity differences. At the same time, it also amplifies high-frequency noise, which reduces stability 
under motion. Laplacian filtering similarly produces a noisier representation, but provides more coherent edge structures than CLAHE, though still fewer reliable features compared to gradient-based operators. In contrast, Sobel and Canny produce sharper and cleaner edge responses with lower noise sensitivity, resulting in more stable feature tracking and improved trajectory estimates.


Table~\ref{tab:hku_vector_all} summarizes results on the HKU and VECtor benchmarks. Edged USLAM achieves an average RMSE of 0.35 m, substantially improving over USLAM~\cite{vidal2018ultimate} (2.64 m) and showing more consistent accuracy than purely event-based pipelines such as PL-EVIO~\cite{PLEVIO}. On the HKU sequences, which emphasize aggressive motion, the proposed method maintains stable performance across all trajectories, whereas USLAM degrades significantly under fast translations and rotations. While event-only or learning-based methods occasionally reach lower errors in specific cases (e.g., PL-EVIO on agg\_tran, DEVO on agg\_rot), they lack robustness across the full set of sequences. Furthermore, event-based methods like DEVO exhibit higher drift than non-event-based methods such as DPVO under insufficient motion, due to the inherent lack of event data.

On the VECtor dataset, which combines both slow and fast motion, Edged USLAM consistently outperforms USLAM and EVI-SAM, particularly in structured low-motion sequences such as sofa\_norm and sofa\_fast, where conventional SLAM pipelines suffer from sparse event information. At the same time, our method remains competitive under fast motion (e.g., robot\_fast) thanks to edge-enhanced event frames and depth priors, which support more reliable landmark initialization. Compared to learning-based baselines (DPVO, DEVO), Edged USLAM offers slightly higher RMSE on average, but has the key advantage of requiring no pretraining or domain adaptation, making it directly applicable to unseen environments in real time.

\subsection{Real-time flight evaluation}

\begin{figure}[!h]
\centering

\begin{minipage}[t]{0.92\linewidth}\centering
  \subfloat[X axis position]{%
    \includegraphics[width=\linewidth]{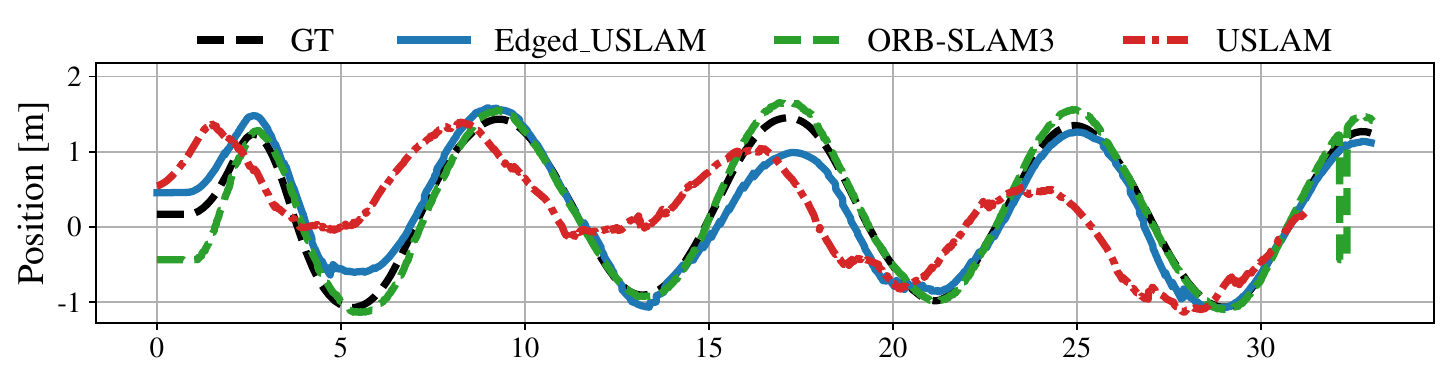}%
    \label{fig:rt_circle_block:x}}\\[0.6ex]
  \subfloat[Y axis position]{%
    \includegraphics[width=\linewidth]{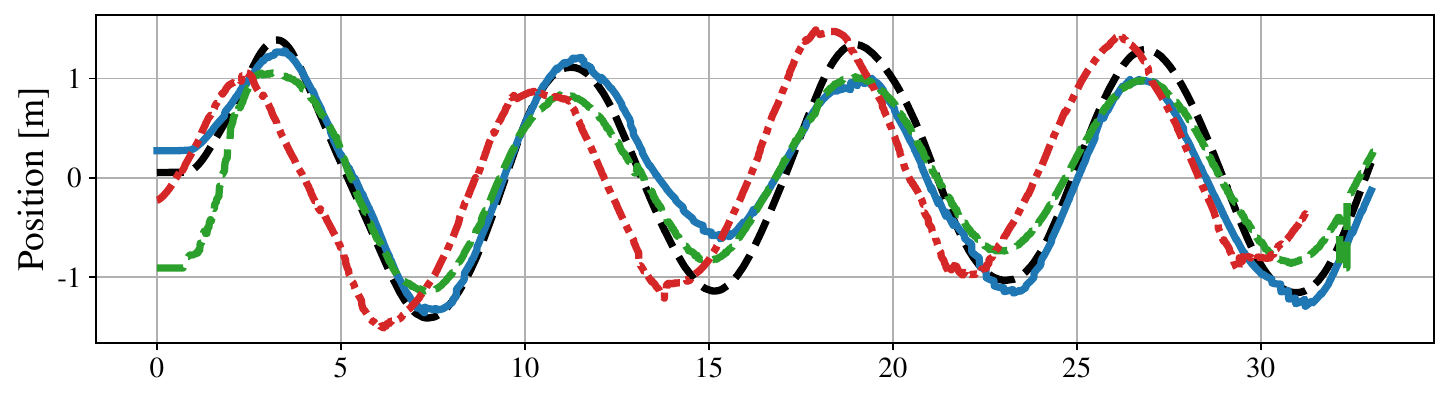}%
    \label{fig:rt_circle_block:y}}\\[0.6ex]
  \subfloat[Z axis position]{%
    \includegraphics[width=\linewidth]{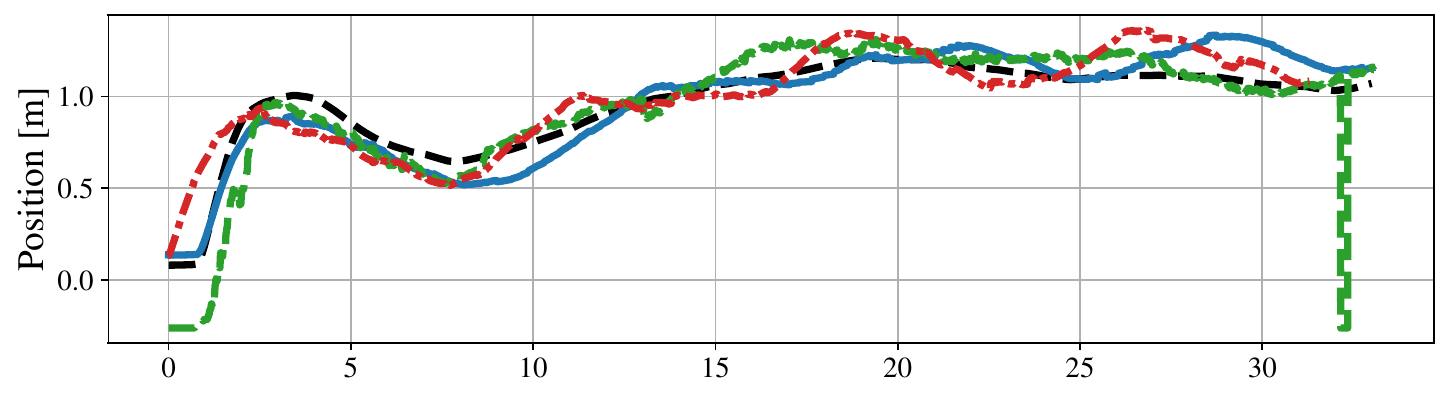}%
    \label{fig:rt_circle_block:z}}\\[0.6ex]
  \subfloat[APE error]{%
    \includegraphics[width=\linewidth]{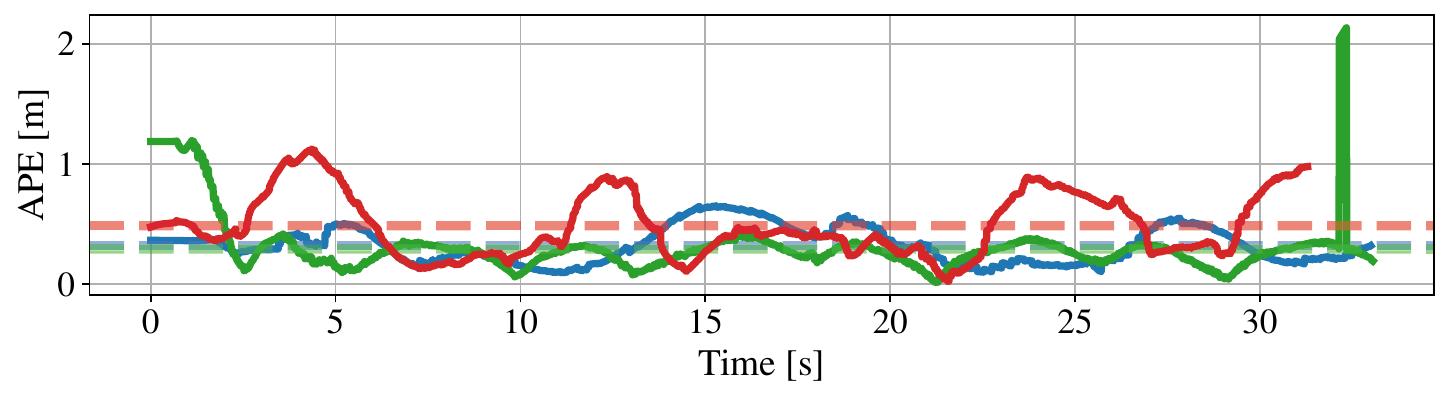}%
    \label{fig:rt_circle_block:ape}}
\end{minipage}

\vspace{1ex}

\begin{minipage}[t]{0.9\linewidth}\centering
  \subfloat[X--Y trajectory]{%
    \includegraphics[width=0.48\linewidth]{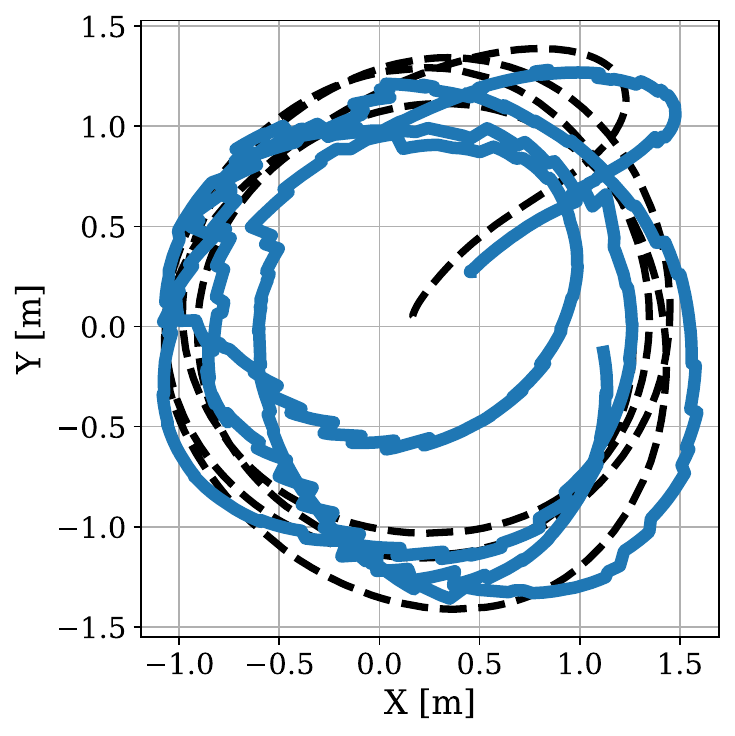}%
    \label{fig:rt_circle_block:xy}}
  \hfill
  \subfloat[Test environment]{%
    \raisebox{4mm}{\includegraphics[width=0.42\linewidth]{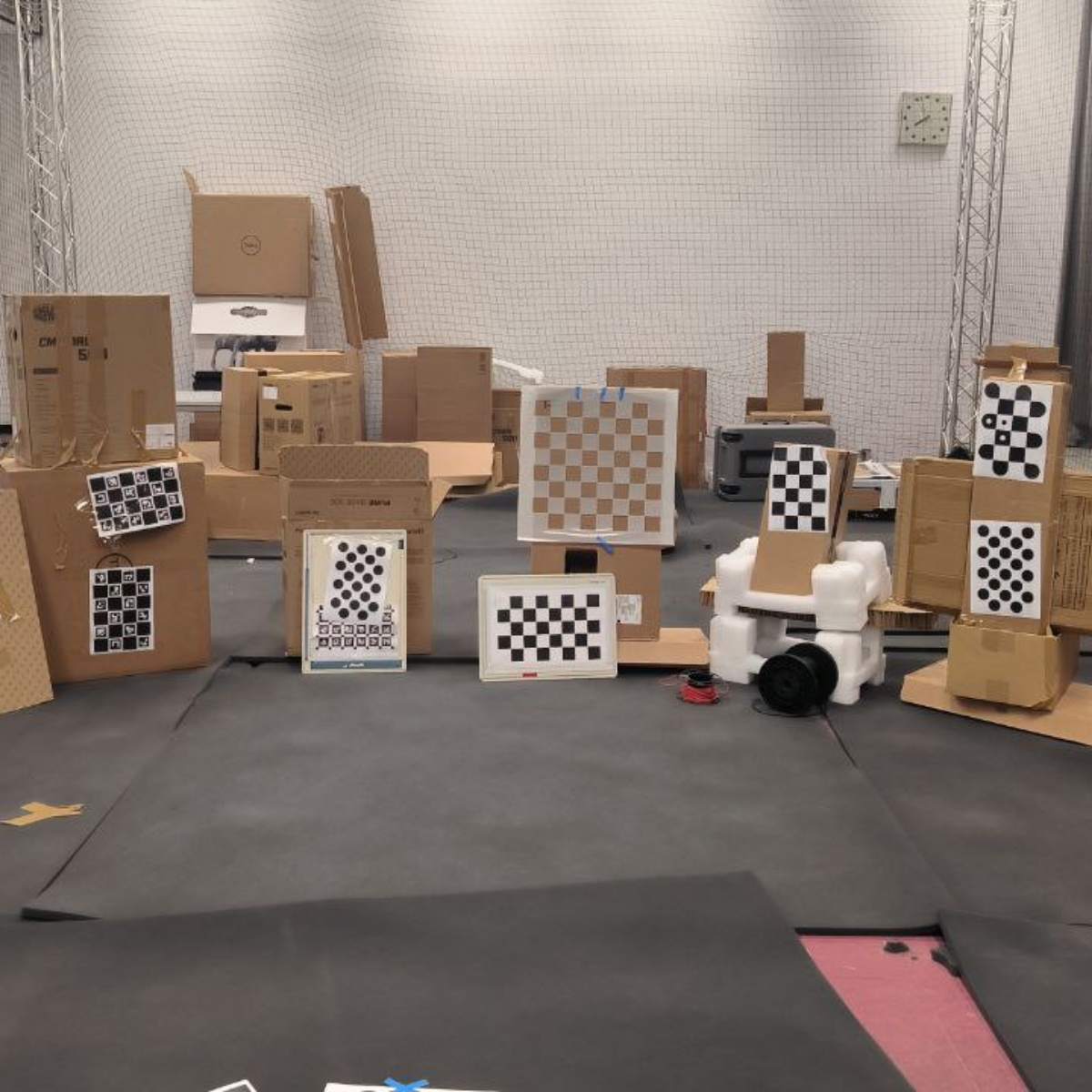}}%
    \label{fig:rt_circle_block:env}}
\end{minipage}

\caption{Real-time circular motion tracking performance of different SLAM methods in well-lit conditions.
Top: estimated axis positions (X, Y, Z) compared with ground truth (black dashed) and absolute position error (APE) over time. 
Bottom: XY trajectory overlay and test environment. 
Mean APE values: Edged USLAM is 0.26\ m, ORB-SLAM3 is 0.21\ m, and USLAM is 0.486\ m. 
}
\label{fig:rt_circle_block}
\end{figure}
All real-time evaluations were conducted on a quadrotor platform in an indoor test arena equipped with a motion capture system for ground-truth trajectory acquisition. The dataset 
collected during these flights covers diverse motion profiles and illumination conditions, including takeoff, landing, and free maneuvers in environments with a realistic number of visual features (neither highly textured nor texture-sparse). The motion profiles include circular flights, aggressive manual trajectories with roll and lateral shifts, low-light aggressive flights, square trajectories through obstacles, straight-line paths, and aggressive yaw rotations. For illumination tests, adjustable room lighting was systematically varied: low light, $30\%$ lit, and $60\%$ lit. Additionally, a dynamic HDR condition was introduced by randomly switching lateral light sources on and off, while a constant HDR setting was created by opening a side window to let sunlight enter from one direction.

Table~\ref{tab:ape_motion_illum} summarizes the APE statistics across these conditions, which are part of our dataset. ORB-SLAM3 achieves strong performance in repetitive scenes (e.g., circle), benefiting from loop closure and back-end optimization. However, its monocular setup introduces significant scale drift, and it fails in low-light and HDR scenarios where standard frames lose texture. It also struggles with aggressive yaw rotations, often losing tracking due to delayed relocalization.
USLAM performs better than ORB-SLAM3 under low illumination and dynamic lighting, thanks to its reliance on events. Nonetheless, it suffers from higher drift and instability during aggressive maneuvers, with APE values frequently exceeding $0.5$–$1.0$ m.
In contrast, the proposed algorithm consistently improves over USLAM across both motion and illumination categories. The edge-enhanced event preprocessing provides more reliable feature generation, resulting in reduced APE (e.g., $0.19$ m under aggressive motion, compared to $0.35$ m with USLAM). Also, Edged USLAM maintains robustness under conditions where ORB-SLAM3 fails (low-light, dynamic HDR). 
Notably, the error under the low-lit condition ($0.34$~m) is lower than in a moderately lit scenario, such as the 30\% lit case ($0.54$~m); this is because under this extreme low-light condition, the system effectively disregards the unreliable RGB stream, whereas under moderate lighting, the integration of noisy visual features introduces residual errors.


Figure~\ref{fig:rt_circle_block} shows the XYZ position estimates and APE analysis for three methods. ORB-SLAM3 achieved the lowest APE ($0.21$\ m), mainly due to its loop closure mechanism. However, it suffered from intermittent tracking failures under fast motion and lighting changes, visible as trajectory discontinuities and APE spikes.
USLAM showed consistent lag and drift along the X-axis (depth), with a higher APE of $0.486$\ m. This was likely caused by delayed pose propagation and limited feature observability.
In contrast, Edged USLAM achieved more stable tracking, particularly along the Y and Z axes, with reduced temporal delay. Its mean APE was $0.26$\ m. While some depth inaccuracies remained, the edge-enhanced front-end allowed better data association and smoother trajectories than USLAM.

%% file: sections/05_Conclusions.tex
\section{Conclusion and Future Work}

The main contribution of this work is Edged USLAM, a hybrid visual–inertial SLAM framework that augments Ultimate SLAM with edge-aware event frame processing and lightweight depth priors. Through evaluations on the Event-Camera, HKU, and VECtor benchmarks as well as real UAV flights, the experiments demonstrate that different SLAM methods excel under different motion and illumination conditions. Event-only pipelines such as PL-EVIO and learning-based methods like DEVO can achieve lower errors in aggressive maneuvers or extreme HDR scenes, highlighting their robustness to fast dynamics and illumination changes. In contrast, Edged USLAM consistently provides more stable tracking and reduced drift in structured and long-duration trajectories, where it outperforms classical baselines including ORB-SLAM3 and USLAM, while maintaining reliable operation in real UAV deployments. The evaluations reveal that while IMU–RGB pipelines remain strongest in slow-motion settings, our edge-aware event processing provides advantages in HDR environments, and the depth-aware event frames improve both accuracy and landmark initialization during aggressive maneuvers and in dynamic environments.

In addition to releasing a drone-based dataset with diverse motion and illumination conditions, future work will focus on improving depth estimation beyond the current lightweight ROI-based design, potentially leveraging event–IMU fusion to reduce dependence on RGB frames. Further enhancements to the back-end optimization and the integration of loop closure are also planned to strengthen global consistency and mapping quality.